\documentclass[sigconf]{acmart}

\AtBeginDocument{%
  }

\copyrightyear{2023} 
\acmYear{2023} 
\setcopyright{acmlicensed}
\acmConference[MM '23] {Proceedings of the 31st ACM International Conference on Multimedia}{October 29--November 3, 2023}{Ottawa, ON, Canada.}
\acmBooktitle{Proceedings of the 31st ACM International Conference on Multimedia (MM '23), October 29--November 3, 2023, Ottawa, ON, Canada}
\acmPrice{15.00}
\acmISBN{979-8-4007-0108-5/23/10} 
\acmDOI{10.1145/3581783.3612444}

\usepackage{multirow}
\begin{document}

\title{FFNeRV: Flow-Guided Frame-Wise Neural Representations for Videos}

\author{Joo Chan Lee}
\affiliation{%
  \institution{Sungkyunkwan University}
  \city{Suwon}
  \country{South Korea}
}
\email{maincold2@skku.edu}

\author{Daniel Rho}
\affiliation{%
  \institution{KT}
  \city{Seoul}
  \country{South Korea}
}
\email{daniel.r@kt.com}

\author{Jong Hwan Ko}
\authornote{Corresponding authors}
\affiliation{%
 \institution{Sungkyunkwan University}
  \city{Suwon}
  \country{South Korea}
}
\email{jhko@skku.edu}

\author{Eunbyung Park}
\authornotemark[1]
\affiliation{%
 \institution{Sungkyunkwan University}
  \city{Suwon}
  \country{South Korea}
}
\email{epark@skku.edu}

\renewcommand{\shortauthors}{Lee et al.}

\begin{abstract}
Neural fields, also known as coordinate-based or implicit neural representations, have shown a remarkable capability of representing, generating, and manipulating various forms of signals. For video representations, however, mapping pixel-wise coordinates to RGB colors has shown relatively low compression performance and slow convergence and inference speed. Frame-wise video representation, which maps a temporal coordinate to its entire frame, has recently emerged as an alternative method to represent videos, improving compression rates and encoding speed. While promising, it has still failed to reach the performance of state-of-the-art video compression algorithms. In this work, we propose FFNeRV, a novel method for incorporating flow information into frame-wise representations to exploit the temporal redundancy across the frames in videos inspired by the standard video codecs. Furthermore, we introduce a fully convolutional architecture, enabled by one-dimensional temporal grids, improving the continuity of spatial features. Experimental results show that FFNeRV yields the best performance for video compression and frame interpolation among the methods using frame-wise representations or neural fields. To reduce the model size even further, we devise a more compact convolutional architecture using the group and pointwise convolutions. With model compression techniques, including quantization-aware training and entropy coding, FFNeRV outperforms widely-used standard video codecs (H.264 and HEVC) and performs on par with state-of-the-art video compression algorithms.
\let\thefootnote\relax\footnotetext{Project page: \href{https://maincold2.github.io/ffnerv/}{https://maincold2.github.io/ffnerv/}}
\begin{figure}[t]
\centering
\includegraphics[width=1.0\columnwidth]{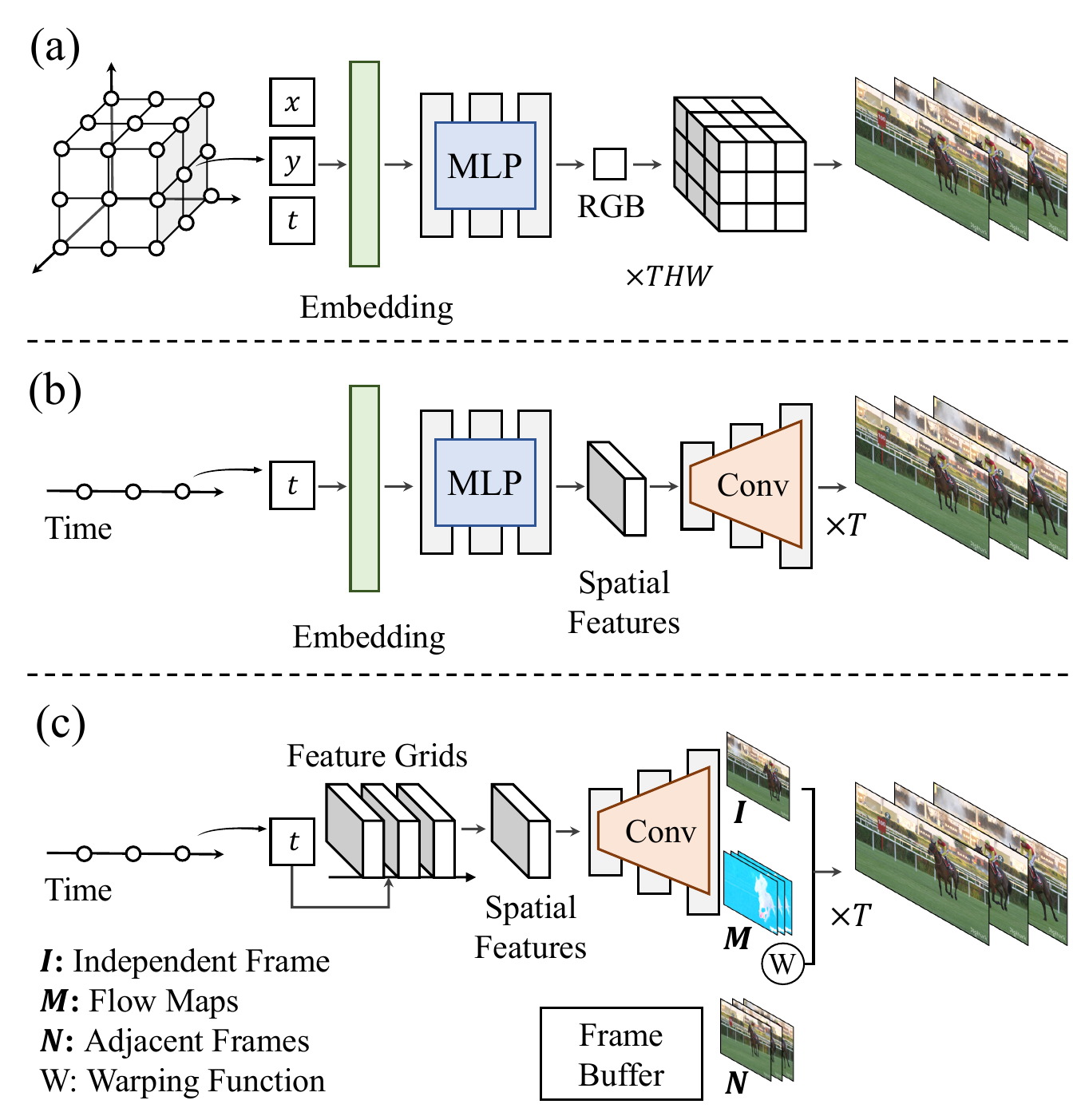}
   \caption{The overall structure of (a) pixel-wise video representations, (b) frame-wise video representations, (c) the proposed flow-guided frame-wise representations (FFNeRV).}
\label{fig_overview}
\end{figure}
\end{abstract}




\maketitle

\section{Introduction}
Recently, research on neural fields~\cite{nf}, which represent signals by mapping coordinates to their quantities (e.g., scalars or vectors) with neural networks, has surged and evoked an increased interest in exploiting their capability in handling various forms of signals, including audio~\cite{SIREN}, image~\cite{liif, mod_siren, streamable}, 3D shape~\cite{loc_grid_3dscene,geo_tree,conv_occu}, and video~\cite{videoinr, video_rho}.
The universal approximation theorem, in concert with coordinate encoding methods, provides the theoretical grounds for accurate signal representation of neural fields~\cite{SIREN,nerf,instant_ngp}.
Recent studies have demonstrated its versatility, beyond simple signal representation, in data compression~\cite{img_comp, video_cohen}, generative models~\cite{gen_img, stylegan5}, and manipulating signals~\cite{editnerf}.

Several works have suggested using this coordinate-based neural representation for data compression, including image and video compression~\cite{coin,img_comp,video_cohen,video_rho} (Figure~\ref{fig_overview} (a)).
Compared to traditional data compression algorithms, such as H.264~\cite{h264} or HEVC~\cite{hevc}, which use highly complex algorithmic pipelines, the process of encoding signals with neural fields (or learning-based methods in general) is much easier to implement, update, and maintain by virtue of the development of machine learning software ecosystems.
Though promising, they fall short of compression performance and, more importantly, require a substantial amount of computations since they require sampling colors for every pixel across all frames.

To overcome the computational bottleneck, \citet{nerv} proposed a frame-wise video representation, coined as NeRV, shown in Figure~\ref{fig_overview} (b).
A stack of MLP and convolutional layers generate a video frame for each time coordinate.
This approach significantly reduced the encoding time compared to the vanilla neural field architecture and performed comparably to the standard video compression algorithms.
The recently proposed E-NeRV \cite{enerv} follows this paradigm while further enhancing video quality. 

We propose flow-guided frame-wise neural representations for videos (FFNeRV), as presented in Figure~\ref{fig_overview} (c) and \ref{fig_arch}.
Inspired by the standard video codecs, we incorporate optical flows into the frame-wise representation to exploit the temporal redundancy.
FFNeRV generates a video frame by aggregating neighboring frames guided by flows, enforcing the reuse of pixels from other frames.
It encourages the network not to waste parameters by memorizing the same pixel values across frames, greatly enhancing parameter efficiency.

Motivated by the grid-based neural representations~\cite{loc_grid_3dscene,plenoxels,tensorf}, we propose to use multi-resolution temporal grids with a fixed spatial resolution to map continuous temporal coordinates to corresponding latent features.
We also propose using a more compact convolutional architecture, which employs group and pointwise convolutions, motivated by lightweight neural networks~\cite{mobilenet,mnasnet,efficientnet} and generative models~\cite{stylegan3}.

Experimental results on UVG dataset~\cite{uvg} show that FFNeRV outperforms other frame-wise methods in both video representation and frame interpolation.
With quantization-aware training and entropy coding, FFNeRV outperforms widely-used video codecs (H.264 and HEVC) and performs comparably with state-of-the-art video compression algorithms.

In summary, our contributions are as follows:
\begin{itemize}
    \item We propose flow-guided frame-wise video representation (FFNeRV) by incorporating optical flows into the frame-wise representation to exploit the temporal redundancy.
    \item Motivated by the grid-based neural representations, we propose to use multi-resolution temporal grids with a fixed spatial resolution to map continuous temporal coordinates to corresponding latent features.
    \item For further efficiency, we devise a compact convolutional architecture and pruning-robust quantization-aware training.
    \item Our approach outperforms others using neural representation or neural fields, while achieving on par with state-of-the-art video compression methods.
\end{itemize}

\section{Related Works}
\subsection{Neural Fields}
Neural fields, also known as implicit neural representations, represent an arbitrary signal by parameterizing neural networks as a function of coordinates.
However, if only low-dimensional coordinates are provided as inputs, neural networks struggle to learn fine details due to their bias to prioritize low-frequency signals~\cite{spectral}.
To successfully learn the high-frequency details, neural fields rely on various methods, such as coordinate encodings (preprocessing)~\cite{nerf,fourier,mipnerf,instant_ngp,VQAD,tensorf} or activation functions~\cite{SIREN,GARF,beyond_periodicity}.
These advancements in neural fields made it possible to show remarkable representation performance in various tasks, including 3D reconstruction \cite{loc_grid_3dscene,conv_occu,geo_tree} and novel view synthesis \cite{nerf,nerfinwild,dir_v_syn}.

Frequency encodings (borrowing the term from Instant-NGP~\cite{instant_ngp}) multiply coordinates with a predefined matrix to map low-dimensional coordinates to higher dimensions followed by sinusoidal functions~\cite{nerf,fourier,mipnerf}.
Another line of approach is parametric encodings, which rely on additional parameters to translate coordinates to latent features, using grids~\cite{loc_grid_3dscene,nsvf,conv_occu,dir_v_syn, plenoxels,instant_ngp,VQAD,tensorf} or octree~\cite{geo_tree,PlenOctrees,mvsoct}.
Although parametric encodings yield improved representation quality and faster convergence compared to frequency encodings, they require additional, often large, memory footprints due to the volumetric structure~\cite{instant_ngp}.
Recent works explored a more parameter-efficient representation with reduced redundancy of dense voxel grids by employing tensor decomposition \cite{tensorf}, hash encoding~\cite{instant_ngp}, and vector quantization~\cite{VQAD}, showing promising performance.
Inspired by the recent success of multiple grids, we propose to adopt temporal grids with multi-resolution for video representation.
The core idea of temporal grids is to achieve an effective plane feature interpolated along only the time axis, with compact representation.

\begin{figure*}[t]
\begin{center}
\includegraphics[width=1.0\linewidth]{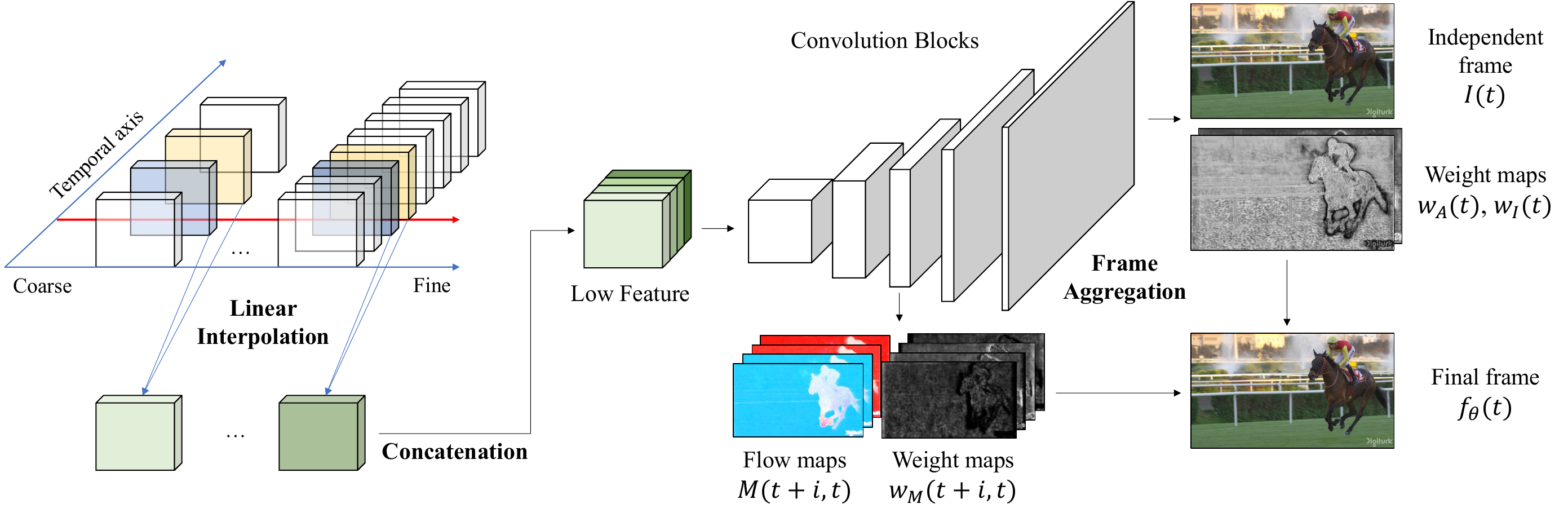}
\end{center}
   \caption{The detailed structure of FFNeRV. For an arbitrary temporal coordinate (red arrowed line), respective planes are obtained from multi-resolution temporal grids. These planes are concatenated to use as inputs for the convolution blocks, which process and upscale latent features and generate flow maps, weight maps, and color maps.}
\label{fig_arch}
\end{figure*}

\subsection{Neural Fields for Video}
Following the success in 3D tasks, the neural field has also been explored for other visual signals, such as images \cite{coin, img_comp, liif} and videos \cite{videoinr, video_cohen, video_rho, scalable}.
The majority of works represent the image and video frames by mapping pixel coordinates to corresponding RGB colors.
However, pixel-wise representations have difficulties representing large and long videos~\cite{nerv}. 
High spatial and temporal resolution hinders the fast training and inference speeds of the pixel-wise representations.

NeRV \cite{nerv} proposed frame-wise video representation, which maps temporal coordinates to their corresponding video frames.
An MLP generates small spatial features at time $t$, and then they are processed and upscaled by following convolutional layers.
NeRV has significantly sped up training and inference times for video representation, enabled by frame-wise sampling.
NeRV demonstrated that frame-wise neural representations can be an efficient way, achieving comparable compression performance to that of standard video codecs.
Based on NeRV, E-NeRV \cite{enerv} proposed a more parameter-efficient architecture for frame-wise representation.
However, these approaches still show limited video representation performance and poor generation of unseen frames.

Inspired by standard video codecs, we propose incorporating optical flows into frame-wise representation to exploit similar patterns across video frames, resulting in a more efficient representation.
To the best of our knowledge, this is the first approach to encoding optical flow maps in frame-wise video representations.
With the temporal grids, our approach outperforms other frame-wise representations for both video representation and frame interpolation.
Furthermore, we introduce improved model compression techniques, resulting in par with state-of-the-art video compression algorithms.

\section{FFNeRV}

\noindent\textbf{Background.} NeRV-like frame-wise video representations express a video as a vector-valued function of time $\mathit{f_\theta}:\mathbb{R} \rightarrow \mathbb{R}^{3 \times H \times W}$, where $H$ and $W$ denote the height and width of the video, respectively.
These representations translate time coordinates to video frames in two stages.
The first stage maps one-dimensional time coordinates to higher dimensional features.
The second stage decodes these features into video frames.
Overall, it can be defined as follows,
\begin{equation}
\label{eq:nerv_like}
    f_\theta(t) := \textrm{NN}_{\textrm{dec}}(\textrm{NN}_{\textrm{feat}}(t)),
\end{equation}
where $\textrm{NN}_{\textrm{feat}}(\cdot)$ and $\textrm{NN}_{\textrm{dec}}(\cdot)$ are the first and second stages of the neural network.
In NeRV~\cite{nerv} and E-NeRV~\cite{enerv}, variants of MLP with positional encoding were used as the first stages.
After the first stage, convolutional layers decodes the output of the first stage in order to obtain the video frames.

\noindent\textbf{Overview.} We present a fully convolutional frame-wise flow representation for videos, as shown in Figure~\ref{fig_arch}.
We use optical flow maps to exploit visual information from neighboring frames.
To this end, given a time coordinate $t$, the fully convolutional decoder, $\textrm{NN}_{\textrm{dec}}(\textrm{NN}_{\textrm{feat}}(t))$, produces a set of five components: $I(t), \{M(t+i,t)\}_{i \in \mathcal{N}}, \{w_M(t+i,t)\}_{i \in \mathcal{N}}, w_A(t), w_I(t)$.
$I(t)\in \mathbb{R}^{3 \times H \times W}$ is a predicted independent frame and $M(t+i,t)\in \mathbb{R}^{H \times W}$ denotes a flow map between $t+i$ and $t$ frames.
$\mathcal{N}$ is a set of indices of neighboring frames, and we empirically found that $\mathcal{N}=\{-2, -1, 1, 2\}$ is a good choice and used it throughout the paper.
$w_M(t+i,t),w_A(t), w_I(t) \in \mathbb{R}^{H \times W}$ are the weights for the process called flow-guided aggregation, which will be further explained in Section~\ref{ssec:flow_guided_agg}.
We generate the final frame through flow-guided aggregation.

In addition to flow representation, we propose to use multiple grids instead of using an MLP to generate low-resolution latent features for a particular temporal coordinate.
Each grid has its own unique temporal resolution so that each grid can cover different temporal granularities.

\subsection{Multi-Resolution Temporal Grid}
Inspired by the recent success of grid-based representations in various visual computing domains, we propose using grids instead of positional encoding and MLP.
Grid $G \in \mathbb{R}^{s \times c \times h \times w}$ is a tensor, where $s$, $c$, $h$, and $w$ are the temporal resolution, the number of channels, height, and width of the grid.
We use a linear interpolation on time dimension to extract the spatial feature $\phi(t,G) \in \mathbb{R}^{c \times h \times w}$ from grid $G$ with the following equation,
\begin{equation}
\label{bi_intp}
\begin{gathered}
    \phi(t,G)=|\hat{t}-n|G[m]+|\hat{t}-m|G[n],\\
    m=\lfloor \hat{t}\rfloor,\, n=\lceil \hat{t} \rceil,
\end{gathered}
\end{equation}
where a temporal coordinate $t$ is normalized to the size of the given $s$, denoted by $\hat{t} = \frac{ts}{T}$ ($T$ is the total number of frames in the video and $t$ is the index of the frame).
$m$ and $n$ are indices of the grid $G$ to be referenced, and $G[x]$ denotes the feature from index $x$ of the grid $G$.
$\lfloor \cdot \rfloor$ and $\lceil \cdot \rceil$ denote floor and ceil operations respectively.

Furthermore, we employ multi-resolution grids with different temporal resolutions but equal spatial resolutions, to make each grid cover its own temporal frequency.
The interpolated outputs from each grid are concatenated to create latent two-dimensional features for given time coordinates.
The extracted feature from multi-resolution grids $\textrm{NN}_{\textrm{feat}}(t) \in \mathbb{R}^{ck \times h \times w}$ can be written as follows.
\begin{equation}
\label{f_t}
    \textrm{NN}_{\textrm{feat}}(t)=\mathsf{concat}(\{ \phi(t,G_k) \}_{k=1}^{K}),\\
\end{equation}
where $k$ is the index of multiple grids and $\mathsf{concat}$ denotes the concatenation of $K$ features along the channel axis.

\begin{figure}[t]
\begin{center}
\includegraphics[width=1.0\linewidth]{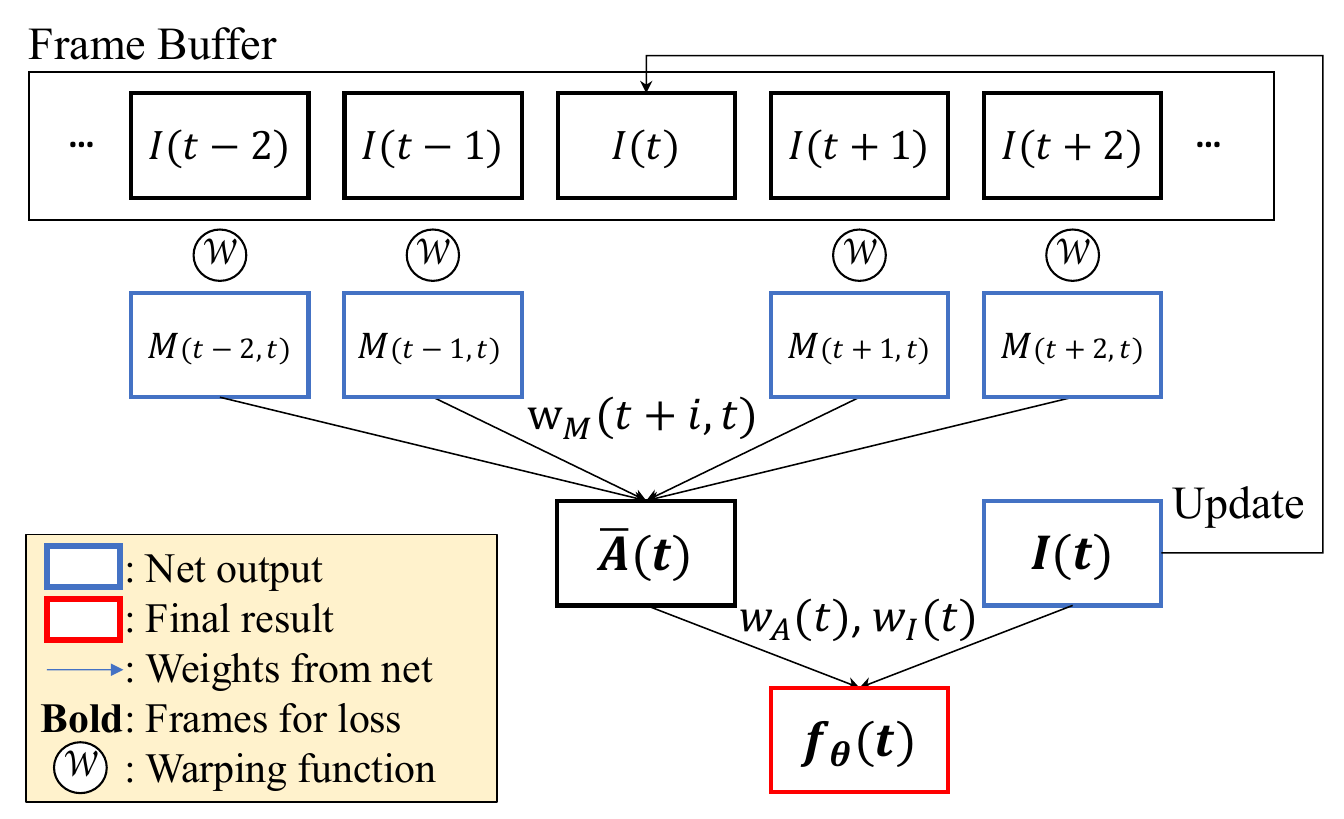}
\end{center}
   \caption{The detailed process of flow-guided frame aggregation.}
\label{fig_agg}
\end{figure}

\subsection{Flow-Guided Frame Aggregation}
\label{ssec:flow_guided_agg}
Optical flows make it possible to exploit similar patterns in adjacent video frames and are thus widely accepted by standard video codecs~\cite{h264,hevc} and neural codecs~\cite{dvc,fvc,hybridcomp}.
We bring this approach to frame-wise video representation by proposing flow-guided frame aggregation.
As described earlier, we use the latent features $\textrm{NN}_{\textrm{feat}}(t)$ from the multi-resolution grids to generate flow maps $\{M(t+i,t)\}_{i \in \mathcal{N}}$ and aggregation weights $\{w_M(t+i,t)\}_{i \in \mathcal{N}},w_A(t), w_I(t)$.
To get the final frame, we warp nearby independent frames with flow maps and combine them with the independent frame at time $t$ using the aggregation weights.
Keep in mind that video frames can already be represented by independent frames.
Although our frame-wise video representation process resembles NeRV and E-NeRV from a broad perspective, it can be distinguished by its explicit exploitation of temporal redundancy. 
Our method takes advantage of visual patterns from surrounding frames, enabled by flow generation and aggregation, significantly improving representation performance.

\noindent\textbf{Frame aggregation.} Figure \ref{fig_agg} illustrates the aggregation process for the final frame construction.
The predicted flow maps warp neighboring independent frames.
Then, we use the weighted sum of the warped frames to get the aggregated frame.
More formally, the aggregated frame $\overline{A}(t)$ is written as,
\begin{small}
\begin{equation}
\label{eq:agg}
\begin{gathered}
    \overline{A}(t) = \sum_{i \in \mathcal{N}}w'_M(t+i,t) \circ \mathsf{warp}({I(t+i)},\,M(t+i,t)), \\
    w'_M(i, t) = \mathsf{exp}(w_M(i, t)) \circ \frac{1}{\sum_{j \in \mathcal{N}} \mathsf{exp}(w_M(t+j,t))},
\end{gathered}
\end{equation}
\end{small}
where $\circ$ denotes element-wise product and $\mathsf{exp}$ represents element-wise exponential function.
We first transform nearby independent frames with the flow maps using bilinear warping.
Then, we apply the softmax function to the weights so that they add up to one for each pixel, and these normalized weights are multiplied with warped frames.
The aggregated frame is the weighted sum of warped frames.

In order to enhance the reconstruction quality, we use the aggregated frame to add details to the independent frame at time $t$.
To this end, we used additional weights $w_A$ and $w_I$ generated by the convolutional decoder.
Similar to frame aggregation weights (Eq.~\ref{eq:agg}), we normalize these two weights ($w_A$, $w_I$) to sum up to one for every pixel.
Thus, the final aggregated frame is defined as follows.
\begin{equation}
\label{eq_f}
\begin{gathered}
    f_\theta(t) = w_A(t) \circ \overline{A}(t) + w_I(t) \circ I(t).
\end{gathered}
\end{equation}

Although standard codecs additionally encode residual parts that are not compensated by reference frames due to a variety of factors (e.g., varied objects/viewpoint or occlusion), FFNeRV does not generate or store residual maps.
We empirically found that the convolution blocks suffer from poor reconstruction quality in non-smooth and noisy residual maps, which results in the final performance degradation.
Instead, our approach fills the residual parts from the current independent frame, enabled by the frame aggregation.
In other words, the independent frame not only contributes to being referenced by neighboring frames but also provides residual terms for the current final frame.

\noindent\textbf{Spatial resolution of outputs.}
The proposed neural network allocates its representational capacity to five different outputs.
For more efficient allocation, we set the spatial resolution of each component differently, as shown in Figure~\ref{fig_arch}.
We set the resolution of independent frames $I(t)$ and weights $w_A$ and $w_I$ to be equivalent to the target video resolution due to their direct involvement in final frame aggregation.
On the other hand, the flow maps $\{M(t+i,t)\}_{i \in \mathcal{N}}$ and the weights $\{w_M(t+i,t)\}_{i \in \mathcal{N}}$ are used in a complimentary way to build the aggregated frame $\overline{A}(t)$. 
Furthermore, as flow maps are typically smoother and less challenging than color maps, they can be efficiently represented with lower resolutions.
Hence, we generate the flow maps and corresponding weights in a low spatial resolution and upscale them to the target video resolution before aggregation.

\noindent\textbf{Training.} Following NeRV, we use the same linear combination of L1 and SSIM loss as the training loss.
Since the quality of the final frame is closely related to the quality of independent and aggregated frames, we include those two frames in the final objective function:
\begin{small}
\begin{equation}
\label{eq_loss}
    \begin{gathered}
    L(\theta) = \frac{1}{T}\sum_{t=1}^{T} \lambda_1 l(\overline{A}(t),F_t) + \lambda_2 l(I(t),F_t) + l(f_\theta(t),F_t), \\
    l(\hat{y},y) = \alpha||\hat{y}-y||_1 + (1-\alpha)(1-\mathsf{SSIM}(\hat{y},y)),
    \end{gathered}
\end{equation}
\end{small}
where $F_t$ denotes the ground truth frame, $L(\theta)$ is the final objective function. 
$\mathsf{SSIM}$ and $||\cdot||_1$ are the structural similarity index measure and L1 norm, respectively.

\noindent\textbf{Frame buffer.}
Computing the training loss requires the independent frames of adjacent video frames.
This incurs an increase in the training time by the length of the frame aggregation window when the loss is naively computed.
To reduce the computational cost, we introduce a frame buffer to store independent frames during the training phase.
Instead of generating independent frames for every training iteration, we draw independent frames from the buffer.
Only the independent frames of the video frames that are sampled as a mini-batch for training will update the buffer.
Figure~\ref{fig_agg} illustrates how the buffer works.
Although this scheme breaks the computational graph for the stored independent frames, the model can parameterize every independent frame by one update per epoch.
As training progresses with the gradually reduced learning rate, the stored independent frames converge to the originals, and finally, the model is learned to generate the optimal final frames without the computational bottleneck.


\begin{figure}[t]
\begin{center}
\includegraphics[width=1.0\linewidth]{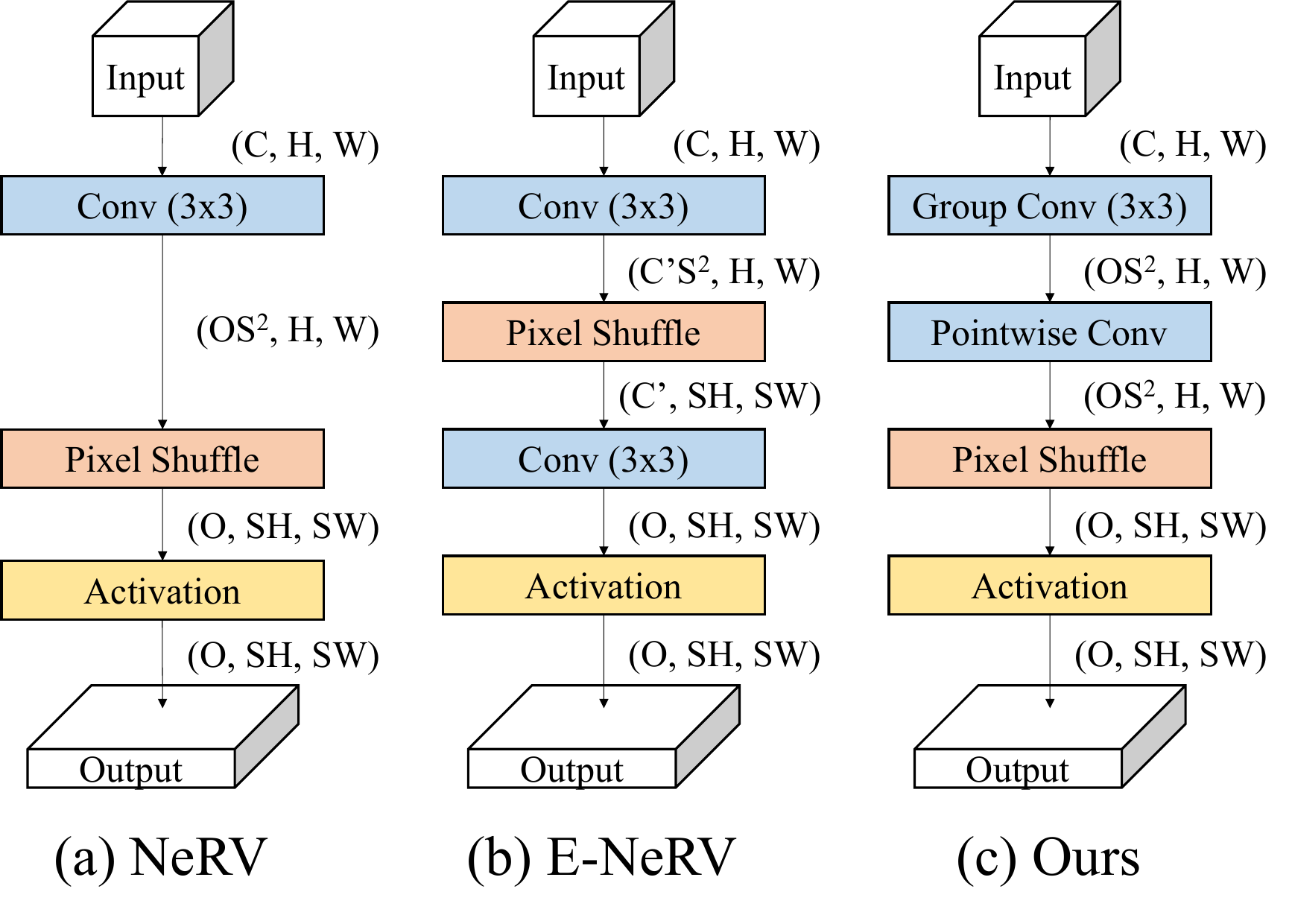}
\end{center}
   \caption{Convolution block architecture of (a) NeRV, (b) E-NeRV, and (c) FFNeRV (ours). C, H, and W are the channel, height, and width of the input feature and O is the output channel. Spatial resolution of the feature is up-scaled by S.}
\label{fig_cb}
\end{figure}

\begin{figure}[t]
\begin{center}
\includegraphics[width=1.0\linewidth]{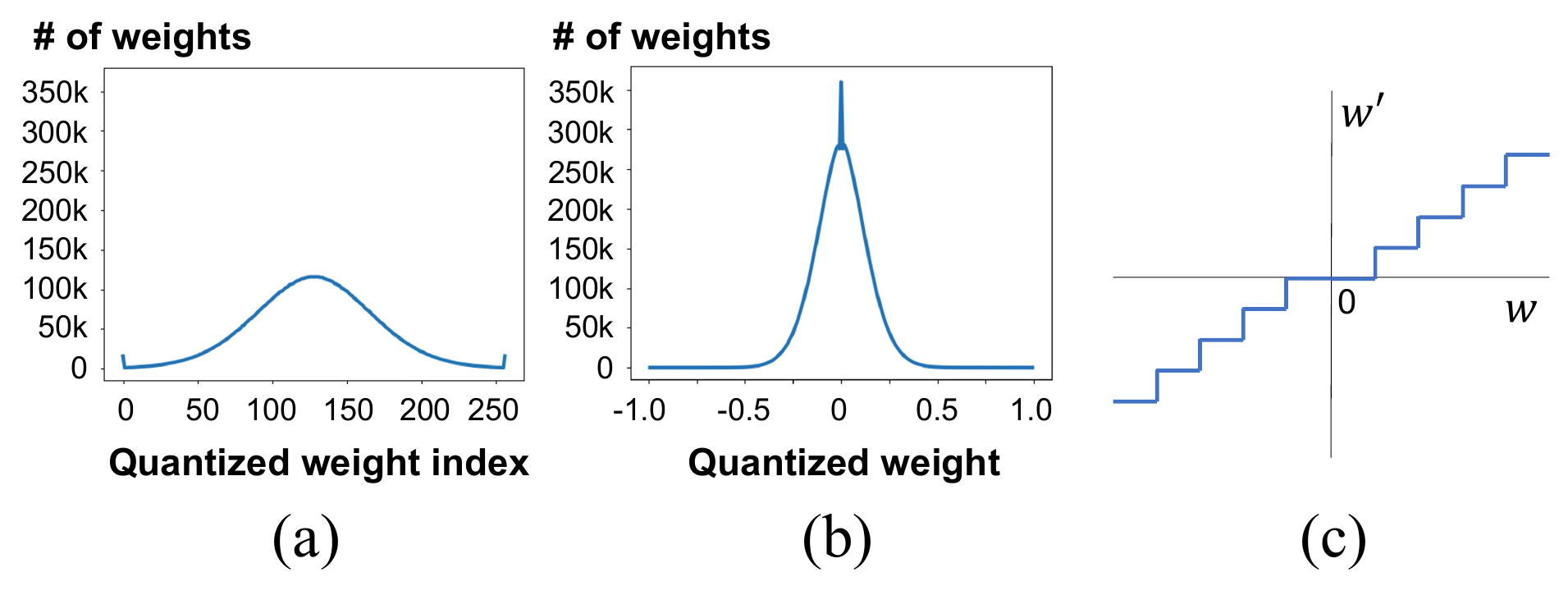}
\end{center}
   \caption{Distribution of all model weights trained for `Jockey' (one of the videos in UVG) quantized by (a) min-max post quantization and (b) QAT.
   Empirical results show that these weight distribution patterns are common, regardless of the video or the network size.
   (c) is the proposed quantization function. 
   }
\label{fig_qat}
\end{figure}

\begin{table*}[t]
\caption{PSNR comparison with other frame-wise video representations on UVG dataset. `Params' denotes the number of model parameters, and `Avg' means an average per video, not per frame. }
\vskip 0.15in
\centering
\begin{tabular}{c|l||ccccccc||c}
\hline
 Method               & Params & Beauty & Bospho & Honey & Jockey & Ready & Shake & Yacht & Avg   \\ \hline\hline
NeRV   & 12.57M          & 35.73           & 37.29           & 41.38          & 36.50           & 29.20          & 36.78          & 31.04          & 35.4  \\ \hline
E-NeRV & 12.49M          & 35.95           & 39.12           & \textbf{41.71}          & 36.89           & 30.61          & \textbf{38.34}          & 32.14          & 36.39 \\ \hline
\textbf{FFNeRV (ours)}   & 12.41M          & \textbf{36.01}           & \textbf{39.78}           & 41.62          & \textbf{38.12}           & \textbf{32.14}          & 37.96          & \textbf{33.64}          & \textbf{37.04} \\ \hline
\end{tabular}
\label{tab_rep}
\end{table*}

\subsection{Model Compression}
Since neural video representations use neural network weights to encode videos, neural network compression is essential in order to achieve high compression performance.
In this section, we describe our proposed compact model architecture and quantization-aware training method.

\noindent\textbf{Compact convolutional architecture.}
To reduce the size of the model itself, we adopt efficient convolution structures.
As pointed out by E-NeRV \cite{enerv}, naively applying spatial convolution with large channels (Figure~\ref{fig_cb} (a)) incurs a number of redundant parameters.
Orthogonal to E-NeRV that injects a channel bottleneck before pixelshuffle operation (Figure~\ref{fig_cb} (b)), we employ group convolution followed by pointwise convolution as shown in Figure~\ref{fig_cb} (c), inspired by depthwise separable convolution \cite{mobilenet}.
When input and output features are with $C_1$ and $C_2$ channels, respectively, spatial convolution with kernel size $k$ produces $C_1 C_2 k^2$ parameters, whereas convolution with $g$ groups followed by pointwise convolution produces $g {C_1\over g} {C_2\over g} k^2 + C_2^2$ parameters.
The compact convolution reduces the number of parameters by a factor of $\left({1\over g}+{C_2 \over k^2 C_1}\right)$, where $C_2$ is not much larger than $C_1$.

\noindent\textbf{Quantization-Aware Training.}
NeRV compresses the representation model in four sequential steps: training the model, model pruning, weight quantization, and weight encoding.
We found that introducing quantization-aware training (QAT)~\cite{dorefa,qat,anypre} can simplify the whole procedure into two or three steps
while improving compression performance.
Quantization-aware training (QAT), as its name implies, ``awares" and reflects quantized weights during training and has been an effective weight quantization technique, which yields fine task performance without additional quantization steps~\cite{dorefa, qat, anypre}.
This technique is based on the straight-through estimator that makes it possible to update weights for non-differentiable operations by passing on non-differentiable values (e.g., quantized weights) in the forward pass but using differentiable values (e.g. non-quantized weights) in the backward pass.
In addition to this, QAT has another advantage for weight encoding.
Figure~\ref{fig_qat} (b) shows the weight distribution under QAT is significantly more concentrated around zero than with NeRV's post min-max quantization (Figure~\ref{fig_qat} (a)).
The dense distribution contributes to improving the efficiency (compression ratio) of entropy coding.
Furthermore, when a large portion of model weights are quantized to zero, the model becomes less sensitive to weight pruning.
Therefore, we introduce pruning-robust $k$-bit quantization enabled by a doubled interval mapped to zero (Figure~\ref{fig_qat} (c)), as following equation,
\begin{equation}
\label{eq_qat}
\begin{split}
    &\textbf{Forward: } w' = \textrm{sign}(w){\lfloor (N\cdot \textrm{tanh}(|w|) \rfloor\over N},\\
    &\textbf{Backward: } {\partial L \over \partial w} \approx {\partial L \over \partial w'},
    \end{split}
\end{equation}
where original weight $w$ is quantized to $w'$, $L$ denotes the objective function and $N = 2^{k-1} - 1$.


\begin{figure*}[ht]
\begin{center}
\includegraphics[width=1.0\linewidth]{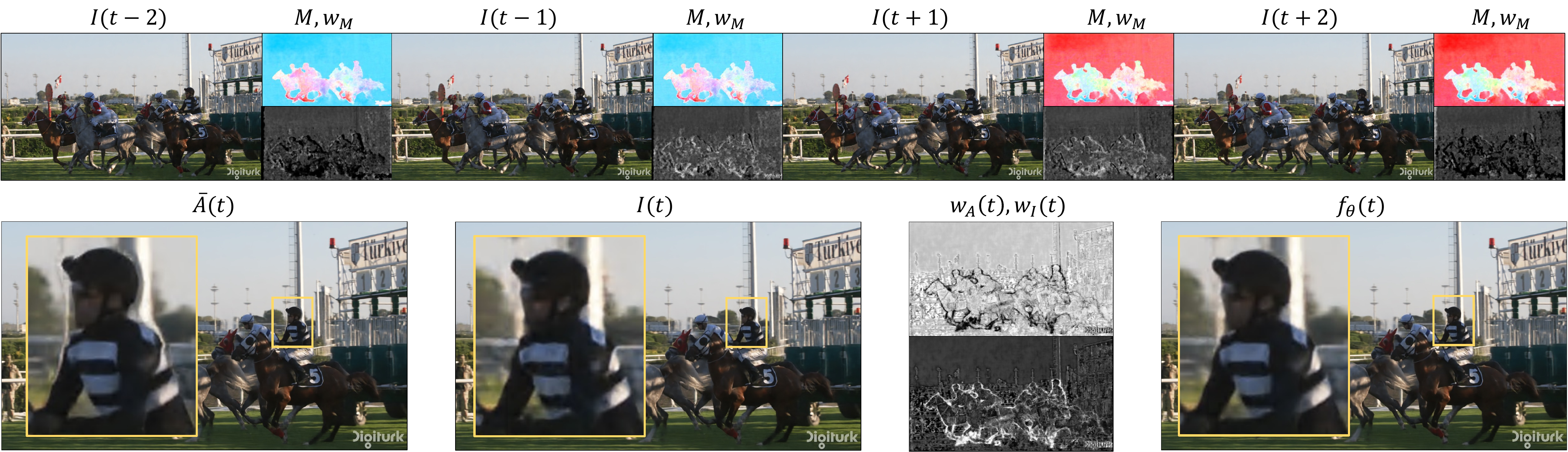}
\end{center}
  \caption{The qualitative results of the proposed method with visualization of frame reconstruction. The whiter pixel in the weight maps indicates a larger value.}
\label{fig_qual}
\end{figure*}

\begin{table*}[]
\caption{PSNR comparison of interpolated unseen frames with other frame-wise video representations using UVG.}
\vskip 0.15in
\centering
\begin{tabular}{c|l||ccccccc||c}
\hline
Method       & Params & Beauty & Bospho & Honey & Jockey & Ready & Shake & Yacht & Avg   \\ \hline\hline
NeRV   & 12.57M          & 27.30           & 29.17           & 39.35          & 19.37           & 16.59          & 29.35          & 22.47          & 26.23 \\ \hline
E-NeRV & 12.49M          & 27.95           & 29.43           & 40.19          & 19.92           & 17.26          & 30.69          & 22.82          & 26.89 \\ \hline
\textbf{FFNeRV (ours)}   & 12.41M          & \textbf{33.09}           & \textbf{38.21}           & \textbf{40.23}          & \textbf{24.17}           & \textbf{26.92}          & \textbf{33.02}          & \textbf{30.73}          & \textbf{32.34} \\ \hline
\end{tabular}
\label{tab_intp}
\end{table*}

\begin{figure*}[ht]
\begin{center}
\includegraphics[width=1.0\linewidth]{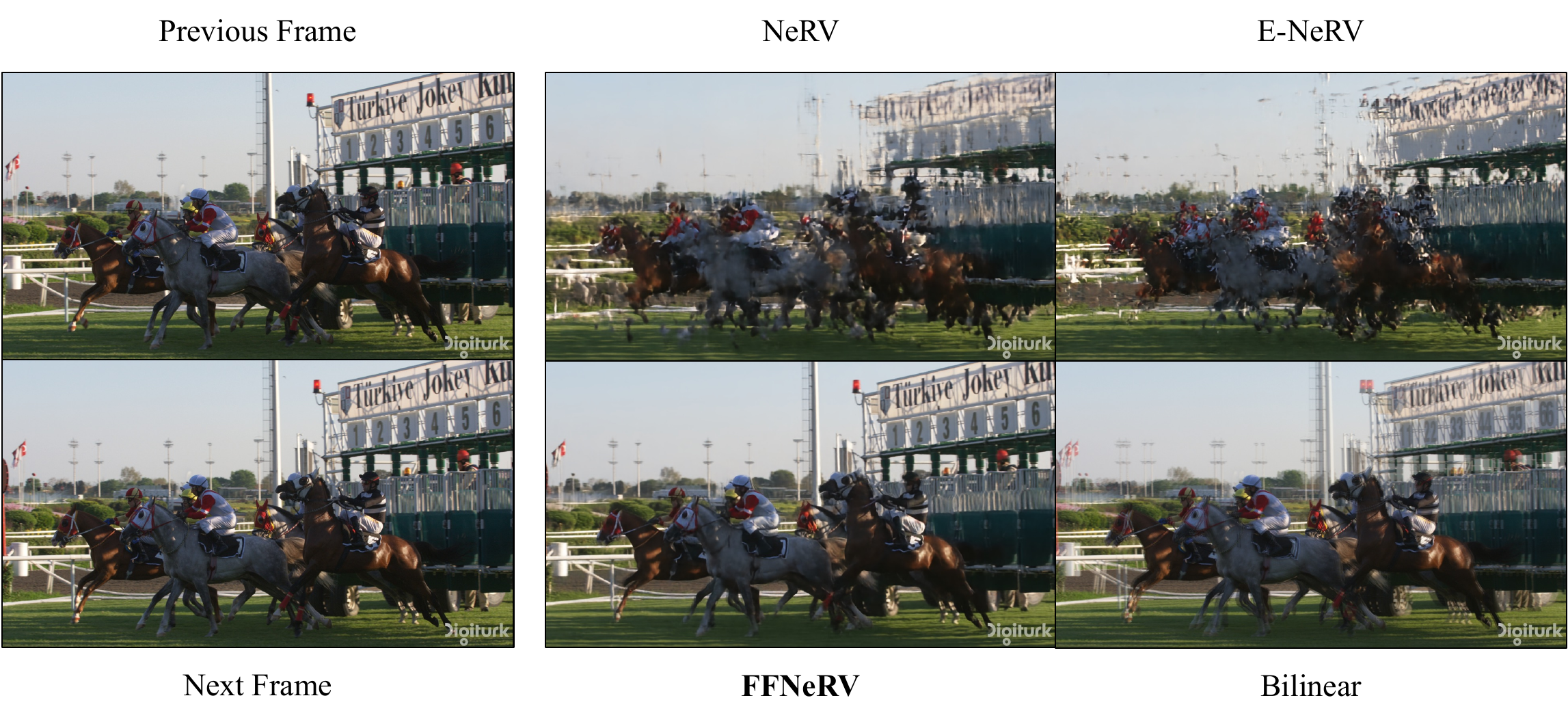}
\end{center}
  \caption{The qualitative results for video interpolation compared with other frame-wise representations and bilinear interpolation.}
\label{fig_qual_intp}
\end{figure*}


\section{Experiments}
\subsection{Experimental Setup}
In this section, we present comprehensive experiments on the UVG dataset \cite{uvg} to validate the effectiveness of FFNeRV.
UVG is commonly used for video compression tasks, containing seven videos and 3900 frames with 1920 × 1080.
In all experiments, $\alpha$, $\lambda_1$, and $\lambda_2$ in Equation \ref{eq_loss} were set to 0.7, 0.1, and 0.1.
We evaluated the quality of video frames by the peak signal-to-noise ratio (PSNR) metric.
We first compared FFNeRV with other frame-wise representation methods for video representation and frame interpolation, and compared with the state-of-the-art algorithms for video compression.
For frame interpolation, we warped the nearest seen frames according to the size of $\mathcal{N}$, to reconstruct a frame at arbitrary time coordinates. 
Given that our model was designed to reconstruct flow maps for fixed distances, we adjusted the flow maps according to the distance between the provided time coordinate and the reference frames. Moreover, we omitted the final aggregation step to enhance visual quality by avoiding the use of the unseen independent frame.
Please refer to the appendix for more implementation details.


\begin{figure}[t]
\begin{center}
\includegraphics[width=1.0\linewidth]{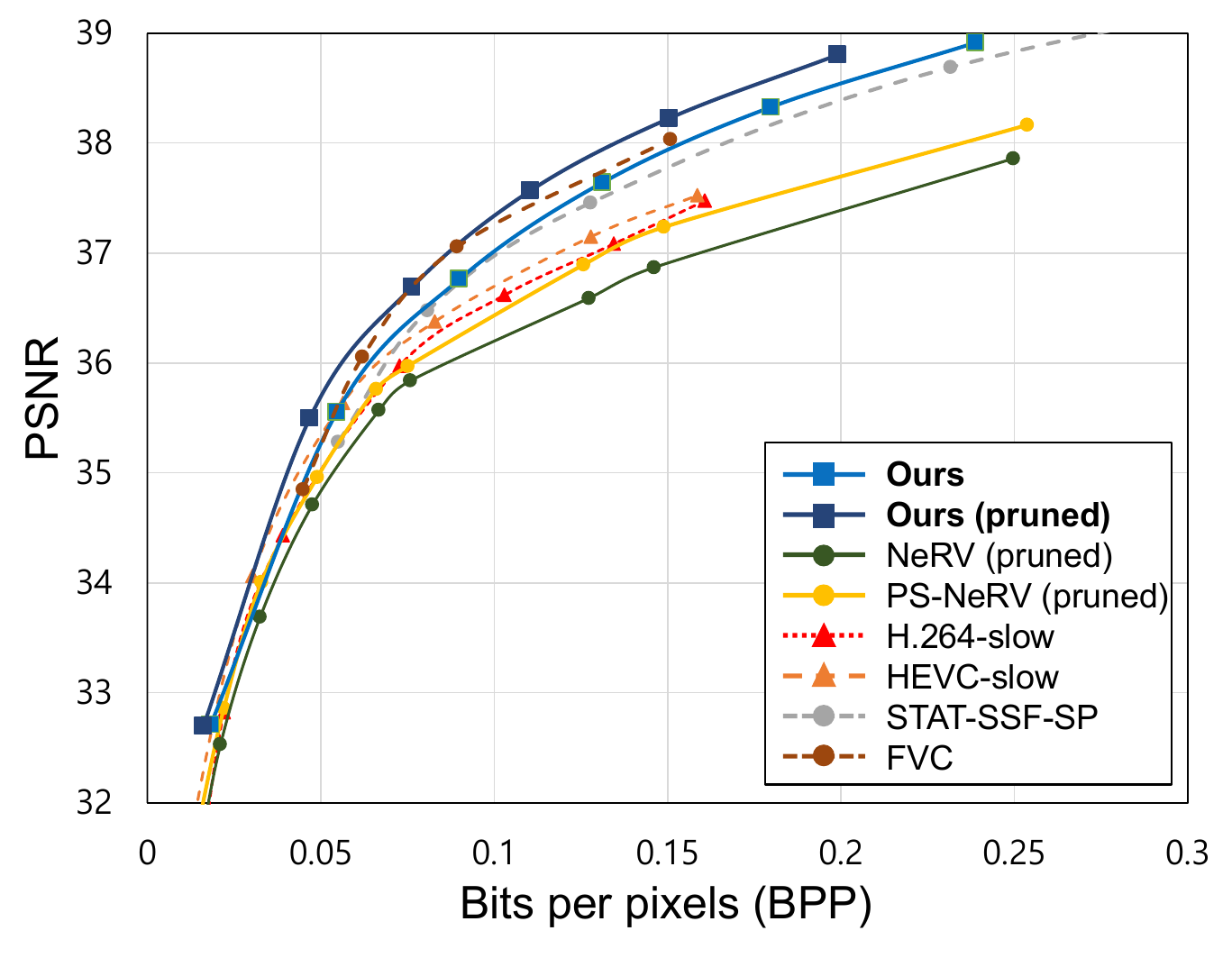}
\end{center}
   \caption{The rate-distortion curve on UVG dataset (best viewed in color).}
\label{fig_comp}
\end{figure}

\subsection{Frame-wise Video Representations}

\noindent\textbf{Video representation.} 
Table~\ref{tab_rep} shows the evaluated video representation performance of frame-wise methods, measured in PSNR.
Although FFNeRV achieves slightly lower PSNR than E-NeRV for relatively static videos such as `Honey' and `Shake', it outperforms baselines by a significant margin overall.
In particular, our approach beats E-NeRV by more than PSNR of 1 in dynamic videos with large motions like `Jockey,' `Ready,' and `Yacht.'
Figure \ref{fig_qual} illustrates the qualitative results of our method with visualization of the frame reconstruction process.
A rapidly moving human torso in the aggregated frame ($\overline{A}(t)$) shows blurred surroundings but clear inside.
To construct a high-quality final frame, the weight map ($w_A(t)$) for the aggregated frame represents low and high values for edges and interior areas, respectively, and vice versa for the independent frame.

\noindent\textbf{Video frame interpolation.}
To measure how well models can predict video frames for unseen continuous time coordinates, we uniformly sampled video frames from each video and used half for training and the other half for validation.
Table \ref{tab_intp} shows PSNRs of the reconstructed unseen frames.
Although E-NeRV yields better performance than NeRV for seen video frames, as shown in Table~\ref{tab_rep}, the average PSNR for unseen frames is nearly identical to NeRV's.
On the other hand, FFNeRV demonstrates superior performance compared to the baselines across all videos, with an average improvement in PSNR of nearly 6 decibels.
Despite the inherent challenges in predicting unseen frames within dynamic videos, such as 'Jockey' and 'Ready,' FFNeRV exhibits considerable enhancement for the overall quality of the predictions.
Figure~\ref{fig_qual_intp} visually compares the FFNeRV's qualitative results against those of NeRV, E-NeRV, and conventional bilinear interpolation, using the 'Ready' video as an example.
Other frame-wise representations struggle to predict dynamic movements, leading to poor visual quality that is even inferior to the naive interpolation. 
In contrast, FFNeRV successfully reconstructs unseen frames with remarkable clarity and high resolution.

\subsection{Video Compression}
This subsection presents the effectiveness of FFNeRV for a major video task, video compression, compared with state-of-the-art methods on the UVG dataset.

\noindent\textbf{Rate-distortion performance.} We compared our approach with widely-used video codecs (H.264, HEVC), neural video representations (NeRV, PS-NeRV \cite{psnerv}), and learning-based models (STAT-SSF-SP \cite{statssfsp}, FVC \cite{fvc}).
H.264 and HEVC were performed with the slow preset mode.
Figure \ref{fig_comp} shows the rate-distortion curve evaluated on the UVG.
Even without pruning, FFNeRV outperforms standard video codecs and neural video representations and performs on par with state-of-the-art video compression algorithms.

\begin{table}[]
\centering
\caption{The decoding speed for a 1080p 600 frames video at 0.1 BPP. $^*$ denotes the reported value in~\cite{fvc_adv}.}
\vskip 0.15in
\begin{tabular}{c|c}\hline
Methods & Decoding FPS (↑) \\\hline\hline
DVC \cite{dvc}     & 3.91             \\
SSF \cite{ssf}    & 3.02         \\
FVC$^*$ \cite{fvc}    & 6.67         \\\hline
H.264 \cite{h264}   & 21.45             \\
HEVC \cite{hevc}   & 19.70             \\\hline
NeRV \cite{nerv}    & 115.01             \\
\textbf{FFNeRV (ours)}  & 84.04             \\\hline
\end{tabular}
\label{tab:dec}
\end{table}

\begin{table}[]
\centering
\caption{The encoding speed and performance evaluated on the UVG dataset. The encoding time reflects the necessary learning duration for each video, while the remaining metrics represent averages calculated across all videos.}
\begin{tabular}{c|c||cc}
\hline
Encoding time  & Method               & BPP (↓) & PSNR (↑) \\ \hline\hline
$\sim$8 hours       & \multirow{2}{*}{NVP} & 0.210   & 36.46    \\
$\sim$11 hours &                      & 0.412   & 37.47    \\ \hline
$\sim$6 hours  & \textbf{FFNeRV}               & \textbf{0.190}   & \textbf{38.05}    \\ \hline
\end{tabular}
\label{tab:enc}
\end{table}

\begin{table*}[t]
\centering
\caption{Ablation studies on the proposed architectures for video representation and frame interpolation.}
\vskip 0.15in

\begin{tabular}{c||c|cc|c||ccccccc||c}
\hline
Task                                                                     & Model & Flow & Grids & Params & Beauty & Bospho & Honey & Jockey & Ready & Shake & Yacht & Avg \\ \hline\hline
\multirow{4}{*}{\begin{tabular}[c]{@{}c@{}}Video\\ Rep.\end{tabular}}    & NeRV           &               &                & 12.57M          & 35.73           & 37.29           & 41.38          & 36.50           & 29.20          & 36.78          & 31.04          & 35.40         \\ \cline{2-13} 
                                                                                  & Flow-only      & \checkmark             &                & 12.57M          & 35.79           & 38.34           & 41.45          & 37.46           & 30.81          & 36.95          & 32.44          & 36.18        \\ \cline{2-13} 
                                                                                  & Grids-only     &               & \checkmark              & 12.41M          & 35.98           & 39.12           & \textbf{41.63}          & 37.38           & 31.00          & 37.77          & 32.71          & 36.51        \\ \cline{2-13} 
                                                                                  & \textbf{FFNeRV}  & \checkmark             & \checkmark              & 12.41M          & \textbf{36.01}           & \textbf{39.78}           & 41.62          & \textbf{38.12}           & \textbf{32.14}          & \textbf{37.96}          & \textbf{33.64}          & \textbf{37.04}        \\ \hline\hline
\multirow{4}{*}{\begin{tabular}[c]{@{}c@{}}Video\\ Interp.\end{tabular}} & NeRV           &               &                & 12.57M          & 27.30           & 29.17           & 39.35          & 19.37           & 16.59          & 29.35          & 22.47          & 26.23        \\ \cline{2-13} 
                                                                                  & Flow-only      & \checkmark             &                & 12.57M          & 31.90           & 37.36           & 40.06          & \textbf{26.00}           & 26.02          & 32.60          & 29.46          & 31.91        \\ \cline{2-13} 
                                                                                  & Grids-only     &               & \checkmark              & 12.41M          & 31.68           & 36.89           & \textbf{40.82}          & 21.97           & 20.25          & 32.15          & 27.30          & 30.15        \\ \cline{2-13} 
                                                                                  & \textbf{FFNeRV}  & \checkmark             & \checkmark              & 12.41M          & \textbf{33.09}           & \textbf{38.21}           & 40.23          & 24.17           & \textbf{26.92}          & \textbf{33.02}          & \textbf{30.73}          & \textbf{32.34}        \\ \hline
\end{tabular}
\label{tab_ab_arch}
\end{table*}

\begin{table}[t]
\caption{Ablation studies on the grid configurations. Grid resolutions indicate the temporal resolutions of multiple grids.}
\begin{tabular}{cc|c||c}
\hline
\multicolumn{2}{c|}{Grid Resolutions}                           & Params                  & PSNR           \\ \hline\hline
\multicolumn{1}{c|}{Single}                    & 240            & \multirow{4}{*}{12.41M} & 37.18          \\ \cline{1-2} \cline{4-4} 
\multicolumn{1}{c|}{\multirow{3}{*}{Multiple}} & 192 288        &                         & 37.52          \\ \cline{2-2} \cline{4-4} 
\multicolumn{1}{c|}{}                          & 96 384         &                         & 37.93          \\ \cline{2-2} \cline{4-4} 
\multicolumn{1}{c|}{}                          & 64 128 256 512 &                         & \textbf{38.12} \\ \hline
\end{tabular}
\label{abl_grid}
\end{table}

\noindent\textbf{Running time.} 
Fast decoding is crucial in many practical scenarios, where videos need to be repeatedly decoded, once they have been encoded (e.g., movies or video sharing platforms).
Table~\ref{tab:dec} shows the decoding time of FFNeRV compared to other methods under a similar memory budget.
All methods were run on a single GPU, except for standard codecs (H.264 and HEVC), which were run on a CPU with 4 threads.
FFNeRV achieves real-time decoding, yielding significantly higher throughput than the standard video codecs without hardware acceleration and learning-based approaches using conventional encoder-decoder architectures, but slightly lower than NeRV due to the additional aggregation process.

Although frame-wise representations encode videos much faster than pixel-wise methods \cite{nerv}, they still fall short of standard codecs or general learning-based methods.
Similar to other frame-wise methods, FFNeRV takes hours of training to encode a 1080p video.
Table~\ref{tab:enc} shows the encoding time and performance of FFNeRV compared to NVP, the state-of-the-art pixel-wise representation, outperforming other methods such as Instant-NGP~\cite{instant_ngp}, Siren~\cite{SIREN}, and FFN~\cite{fourier}.
FFNeRV outperforms NVP in terms of compression efficiency under a shorter encoding duration.
Although promising, the encoding duration remains a limitation when comparing our method to standard codecs or neural codecs. 
Improving the training speed of neural fields is an active research area and we believe it can be further reduced in future works. 

\noindent\textbf{Summary.} 
FFNeRV demonstrates not only promising performance but also significantly fast decoding speed. 
In addition, it requires only a trained network for decoding, resulting in high flexibility and easy deployment, as opposed to the standard codecs requiring specific designs for implementation and deployment. 
Although H.264 can achieve very low latency under hardware/software support, more recent codecs, such as H.266, still struggle to be widely deployed due to their increased complexity. 
On the other hand, learning-based compression methods using encoder-decoder neural networks allow for a straightforward implementation with high compression performance.
However, they are often limited in many practical scenarios due to the need for a large number of training examples and poor decoding speed. 
We believe FFNeRV can be a promising alternative for real-world video compression thanks to its high compression performance, fast decoding, and simple deployment.

\subsection{Ablation Studies}

\noindent\textbf{Model architecture.}
We have made two primary architectural proposals: flow-guided aggregation and multi-resolution temporal grids.
Table~\ref{tab_ab_arch} shows how each proposal contributes to video representation and frame interpolation performance.
We started with NeRV, which uses positional encoding and MLP to encode a latent feature, followed by convolutional blocks that produce the independent frame. 
Flow-only's convolutional blocks generate the independent frame and flow maps, which are aggregated into the final frame based on the latent feature (from positional encoding and MLP). 
Grids-only substitutes the positional encoding and MLP by the proposed temporal grids to encode the latent feature while only generating the independent frame.

For video representation, both architectures enhance the performance orthogonally, with the combined model achieving the highest PSNR.
For predicting unseen frames, flow-guided aggregation largely contributes to high performance, even in instances where the model solely employing multiple grids already outperforms NeRV. 
For instance, in the case of the ‘Ready,’ which contains many fast motions, the PSNR increased by more than 6 decibels compared to when only the grids are incorporated, proving the effectiveness of the flow guidance.

To validate the efficacy of employing multiple temporal grids, we evaluated the performance under various grid configurations. 
Table~\ref{abl_grid} shows the experimental results evaluated on a video, ‘Jockey’. 
Under the same number of parameters, multi-resolution grids yield superior results compared to single-resolution grids, and a higher maximum resolution contributes to improved performance.


\begin{figure}[t]
\includegraphics[width=1.0\linewidth]{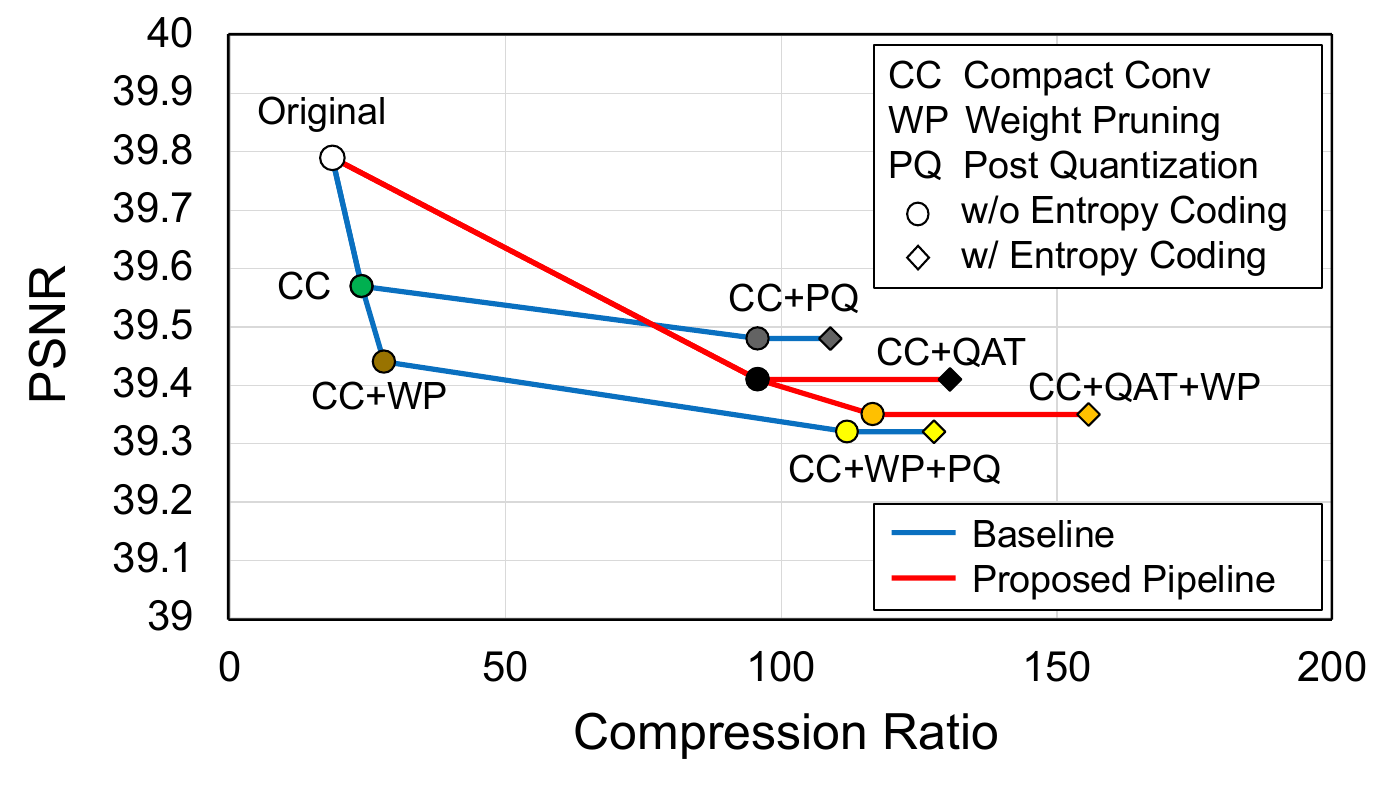}
\caption{Ablation studies on the proposed model compression techniques. Markers of the  same color indicate the same trained model.}
\label{fig_abl_comp}
\end{figure}

\noindent\textbf{Model compression methods.}
We conducted an ablation study on the proposed compression pipeline including the compact convolution and pruning-robust QAT.
Both the grid values and convolutional weights were compressed by the 8-bit width QAT and Huffman encoding~\cite{huffman}.
On the other hand, when applying weight pruning, we did not prune the grid values in order to prevent losing the spatial information of latent features, while eliminating 20\% of the total convolutional weights and additionally trained for 100 epochs.

Figure \ref{fig_abl_comp} depicts compression ratio enhancements caused by each proposed technique, compared to the pipeline implemented in NeRV.
The compact convolution block eliminates almost a quarter of the model parameters with a tolerable accuracy drop.
Compared to post quantization, QAT permits the removal of an additional quantization step with negligible accuracy loss. 
Moreover, this loss is compensated by the higher entropy coding efficiency due to the concentrated quantized weights.
This appears to be the case when the weights are pruned.
Furthermore, the proposed QAT halves the accuracy drop caused by the following pruning, demonstrating its robustness against pruning enabled by the doubled zero interval.
After that, the proposed pipeline yields a higher accuracy and compression ratio with a simpler pipeline compared to the baseline.

\section{Conclusions}
We have proposed a novel frame-wise video representation, FFNeRV, which incorporates frame-wise flow guidance and multi-resolution temporal grids.
Experimental results show FFNeRV outperforms the existing frame-wise methods by a significant margin.
Furthermore, FFNeRV has the capability to reconstruct video frames at arbitrary temporal coordinates, maintaining high quality even in dynamic videos, distinct from other frame-wise approaches.
With the proposed model compression techniques, including an efficient convolutional architecture and pruning-robust QAT, FFNeRV performs on par with state-of-the-art video compression algorithms.
We have demonstrated the practical effectiveness of FFNeRV in video compression, considering compression efficiency, decoding speed, and deployment.
These results are optimistic for the advancement of neural representation-based video compression in the future.

\section*{Acknowledgements}
This work was supported by the Ministry of Science and ICT (MSIT) of Korea under the National Research Foundation (NRF) grants (2022R1F1A1064184, 2022R1A4A3032913) and Institute of Information and Communication Technology Planning Evaluation (IITP) grants (IITP-2019-0-00421, IITP-2023-2020-0-01821, IITP-2021-0-02052, IITP-2021-0-02068), and by the Technology Innovation Program (RS-2023-00235718) funded by the Ministry of Trade, Industry \& Energy (1415187474).


\bibliographystyle{ACM-Reference-Format}
\balance
\bibliography{references}

\appendix

\section{Implementation Details}
\label{app_imp}
We implemented FFNeRV based on NeRV \cite{nerv} official codes in the Pytorch framework, and we used the NVIDIA A100 GPU for all evaluations.
All the results were evaluated once with the fixed random seed.

\subsection{Dataset Preprocessing}
We used the UVG dataset \cite{uvg}, which is commonly used for video compression tasks.
UVG contains six 600 frame videos: `Beauty', `Bosphorus', 'HoneyBee', `Jockey', `ReadySetGo', and `YachtRide', and a 300 frame video: `ShakeNDry', with a resolution of 1920 $\times$ 1080.
We extracted frames from the original YUV videos using ffmpeg \cite{ffmpeg} with the following command:

$\mathsf{ffmpeg\,\; \text{-}i\,\;} video\mathsf{.y4m\,\;} video\mathsf{/f\%05d.png}$

\noindent where $video$ is the file name of each video.

\subsection{Detailed Model Architecture}
\noindent\textbf{Comparison with frame-wise methods.}
As the experimental settings in E-NeRV \cite{enerv},
all video frames were resized to a resolution of 1280 × 720 and the up-scale factors of each convolution block were set to 5, 2, 2, 2, 2 to reconstruct 1280 × 720 frames from the feature map of size 16 × 9.
We evaluated models using every frame in each video, as opposed to the experimental setting in E-NeRV that samples 150 frames from seven videos each.
The performance of baselines was evaluated using the same experimental setting but with their official codes.
We trained models on each video using the adam optimizer for 300 epochs with a batch size of 1.
To configure our model with a similar number of parameters to the baselines, we set the temporal resolution of multi-resolution grids to 64, 128, 256, and 512 for sequences of 600 frame videos and 192, 288 for sequences of 300 frame videos.
Table \ref{fig_fw} shows the detailed model architecture used to compare our approach with other frame-wise representations.
\begin{table}[h]
\centering
\caption{Detailed model architecture to compare with other frame-wise representations.}
\vskip 0.1in
\begin{tabular}{c|c|c}
\hline
Layer & Modules                    & Output Size (C $\times$ H $\times$ W)    \\ \hline\hline
0     & Multi-resolution grids     & 156 $\times$ 16 $\times$ 9                 \\
1     & Convolution block                 & 156 $\times$ 80$\times$ 45                \\
2     & Convolution block              & 96 $\times$ 160$\times$ 90    \\
3     & Convolution block              & 96 $\times$ 320$\times$ 180    \\
3'    & Head layer (for $M$, $w_M$)     & 12$\times$ 320$\times$ 180             \\
4     & Convolution block              & 96 $\times$ 640$\times$ 360    \\
5     & Convolution block              & 96 $\times$ 1280$\times$ 720 \\
5'    & Head layer (for $I$, $w_A$, $w_I$) & 5$\times$ 1280$\times$ 720  \\
& Aggregation & 3$\times$ 1280$\times$ 720 \\
\hline
\end{tabular}
\label{fig_fw}
\end{table}

\noindent\textbf{Video compression.}
We used UVG videos with the original resolution (1920 × 1080) to compare with state-of-the-art methods on the video compression task.
We set the up-scale factors of convolution blocks to 5, 3, 2, 2, 2 for 1920 × 1080 frames.
We trained respective models for seven videos, with temporal grid resolutions of 300, 600 for 600 frame videos, and 150, 300 for 300 frame videos.
We applied the compact convolution blocks except for the first block, and set group of the convolution to 8.
Table \ref{tab_co} depicts the detailed model architecture.

\begin{table}[ht]
\centering
\caption{Detailed model architecture for video compression.}
\vskip 0.1in
\begin{tabular}{c|c|c}
\hline
Layer & Modules                    & Output Size (C $\times$ H $\times$ W)    \\ \hline\hline
0     & Multi-resolution grids     & $C_1$ $\times$ 16 $\times$ 9                 \\
1     & Conv block                 & $C_2$ $\times$ 80$\times$ 45                \\
2     & Compact Conv block              & max($C_2$/2, S)$\times$ 240$\times$ 135    \\
3     & Compact Conv block              & max($C_2$/4, S)$\times$ 480$\times$ 270    \\
3'    & Head layer (for $M$, $w_M$)     & 12$\times$ 480$\times$ 270             \\
4     & Compact Conv block              & max($C_2$/8, S)$\times$ 960$\times$ 540    \\
5     & Compact Conv block              & max($C_2$/16, S)$\times$ 1920$\times$ 1080 \\
5'    & Head layer (for $I$, $w_A$, $w_I$) & 5$\times$ 1920$\times$ 1080  \\       
& Aggregation & 3$\times$ 1920$\times$ 1080 \\
\hline
\end{tabular}
\label{tab_co}
\end{table}

We evaluated models with different sizes by changing the value of $C_1$, $C_2$, and $S$.
These values were set differently depending on the length of each video, in order to match BPP simlilar at every level.
They were set to (24, 192, 16), (48, 392, 24), (64, 512, 32), (80, 640, 48), (96, 768, 48), and (112, 896, 54) for 600 frame videos and (16, 128, 16), (32, 256, 24), (48, 384, 32), (56, 448, 48), (64, 512, 48), and (80, 640, 54) for 300 frame videos.

\begin{figure}[ht]
\begin{center}
\includegraphics[width=1.0\linewidth]{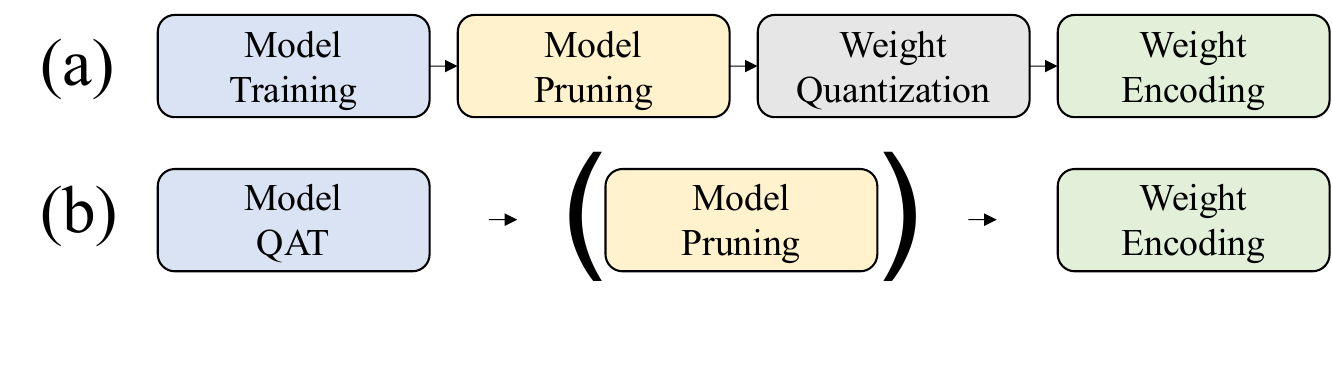}
\end{center}
   \caption{Video compression pipeline of (a) NeRV and (b) Ours. The parenthesis surrounding ''model pruning" in (b) means that it can be skipped.}
\label{fig_cpip}
\end{figure}

Since neural video representations use neural network weights to encode videos, it is necessary to compress neural networks in order to achieve high compression performance.
As described in Figure \ref{fig_cpip} (a), NeRV compresses the representation model in four sequential steps.
On the other hand, we simplify the whole procedure into two or three by applying QAT that reflects quantized weights during training, as depicted in Figure~\ref{fig_cpip} (b).
We applied QAT with 8-bit width and Huffman encoding  \cite{huffman}. Only with the two steps, compression performances of each video at various levels are reported in Table \ref{tab_comp_res}.

NeRV employed unstructured weight pruning, which may produce a non-negligible storage cost for the indices that specify which components of the network weights in some computer systems~\cite{weiprune}.
Therefore, we evaluated our models with and without unstructured pruning, respectively.
When applying weight pruning, we eliminated 20\% of the total convolutional weights and additionally trained for 100 epochs.


\begin{figure}[ht]
\begin{center}
\includegraphics[width=1.0\linewidth]{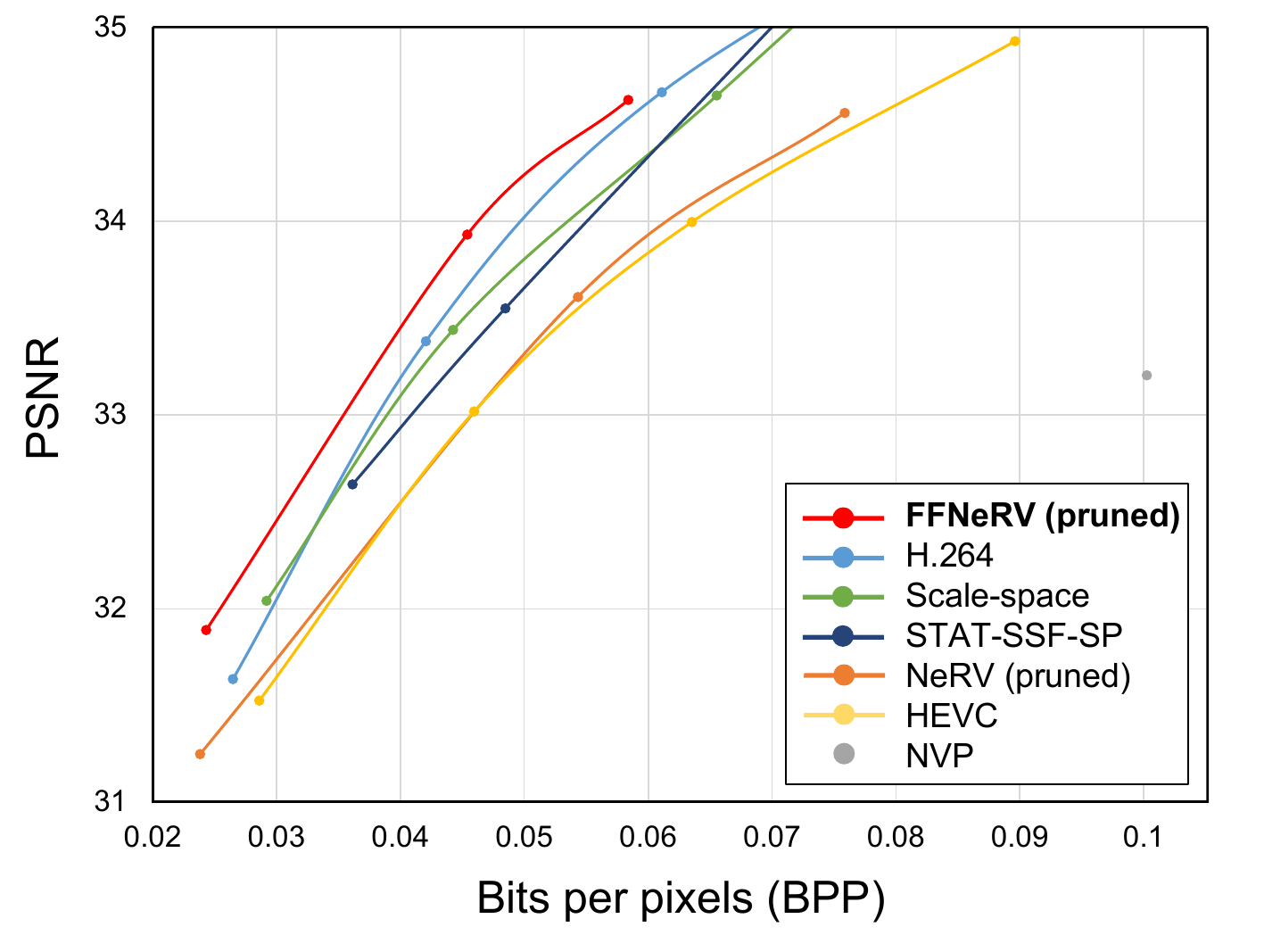}
\end{center}
   \caption{The rate-distortion curve on MCL-JCV dataset (best viewed in color).}
\label{fig_comp_mcl}
\end{figure}

\section{More diverse videos}
We performed additional experiments on a widely-used video dataset, MCL-JCV~\cite{mcl}, which contains thirty 1080p videos at 30fps. 
Compared to UVG, scenes and camera movements are much more dynamic.
We compare FFNeRV with other video codecs, which include standard codecs (H.264~\cite{h264}, HEVC~\cite{hevc}), learning-based methods (Scale-space~\cite{scalespace}, STAT-SSF-SP~\cite{statssfsp}), and neural representations (NeRV~\cite{nerv}, NVP~\cite{scalable}).
As shown in Figure~\ref{fig_comp_mcl}, the proposed FFNeRV consistently outperforms other baselines.

\section{Additional Ablation Studies}
We conducted additional ablation studies using the model trained for `Jockey', whose $C_1$, $C_2$, and S are set to (48, 392, 24).

\begin{table}[t]
\centering
\caption{Ablation studies on output spatial resolution and residual terms.}
\vskip 0.15in
\begin{tabular}{cc|c||c|c}
\hline
\begin{tabular}[c]{@{}c@{}}Low\\ resolution\end{tabular} & \begin{tabular}[c]{@{}c@{}}High\\ resolution\end{tabular} & Residuals & 32bit & \begin{tabular}[c]{@{}c@{}}8bit\\ (QAT)\end{tabular} \\ \hline\hline
-                                                        & All &                                                       & 37.40 & 36.61      \\ \hline
$M$, $w_M$                                                  & $I$,  $w_A$, $w_I$                                            && 37.49 & \textbf{37.29}     \\ \hline
$M$, $w_M$                                                  & $I$,  $w_A$, $w_I$                                            &\checkmark& 37.55 & 37.27     \\ \hline
\end{tabular}
\label{tab_abl_spa}
\end{table}

\subsection{Output Spatial Resolution}
Table~\ref{tab_abl_spa} shows the effect of using different spatial resolutions for each output component. 
Using different spatial resolutions or not has no noticeable effect on the performance of full-precision (32-bit) models.
We interpret this is due to high representation capacity of models.
However, the performance gap gets wider as the bit precision of the network decreases, and setting every output component's resolution to high resolution appears to come at a cost.
We improved the video representation performance of low-precision networks by allocating different spatial resolutions in accordance with their importance.

\subsection{Residual Terms}
Although hand-crafted codecs utilize not only optical flows but also residuals, we validated the effectiveness of the flow representation without residuals in the paper.
Table \ref{tab_abl_spa} compares the results with and without residual terms.
With residuals, the final frame is defined as,
\begin{equation}
\label{eq_f}
\begin{gathered}
    f_\theta(t) = w_A(t) \circ \overline{A}(t) + w_I(t) \circ I(t) + R(t),
\end{gathered}
\end{equation}
where $R(t)$ is the residual map at time $t$.
When the model has full-bit precision, residuals boost performance marginally.
In contrast, residual somewhat degrades the 8-bit model's performance.
As we are primarily concerned with efficient video representation, we omit the residual terms.
The results indicate that the independent frame efficiently contributes to filling residual parts, enabled by flow aggregation, as mentioned in the paper.

\begin{table}[ht]
\centering
\caption{PSNR with different batch size.}
\vskip 0.15in
\begin{tabular}{c|c||c}
\hline
Batch size & Initial LR & PSNR  \\ \hline
1          & 0.0005     & 37.29 \\ \hline
2          & 0.001      & 37.49 \\ \hline
4          & 0.002      & 37.56 \\ \hline
8          & 0.004      & not converge\\ \hline
\end{tabular}
\label{tab_abl_bs}
\end{table}

\subsection{Batch Size}
We evaluated the performance with different batch sizes, setting the initial learning rate to increase as the batch size, as shown in Table \ref{tab_abl_bs}.
The larger batch size results in higher performance, except for the case of non-convergence.
Although we set the batch size to 1 throughout all experiments, these results show that our model's performance can be enhanced by only adjusting the batch size and initial learning rate.

\section{Additional Qualitative Results}
\begin{figure*}[ht]
\begin{center}
\includegraphics[width=1.0\linewidth]{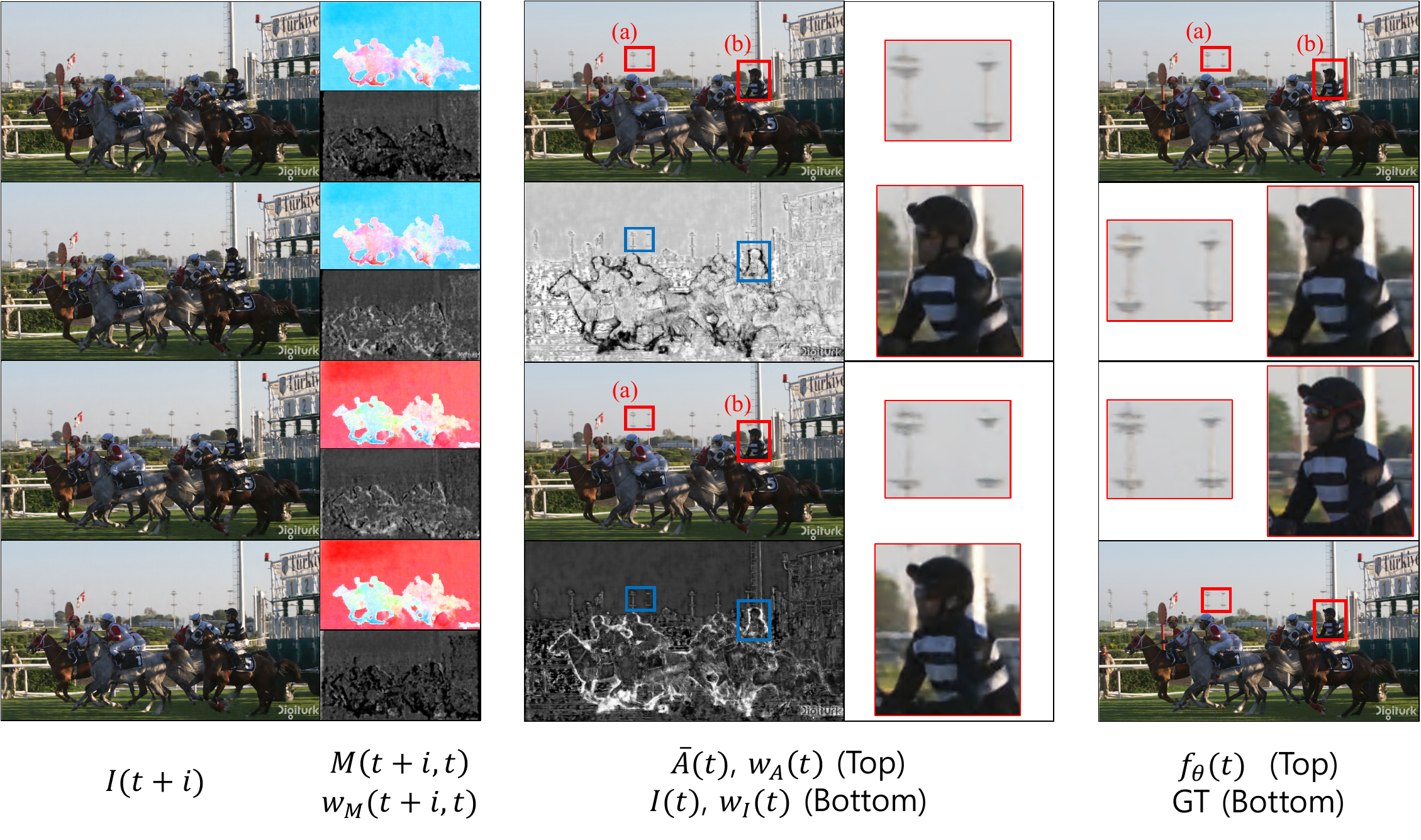}
\end{center}
   \caption{The qualitative results of the proposed method with visualization of frame reconstruction. Two parts of the frames (a) and (b) are emphasized for better visualization. The whiter pixel in the weight maps indicates a larger value.}
\label{fig_sup_qual1}
\end{figure*}

Figure \ref{fig_sup_qual1} shows a visualization of the frame reconstruction process together with the qualitative outcomes of FFNeRV.
The square area (a) in the figure is the background and constantly appears in the following frames.
Although the independent frame $I(t)$ misses this faint part, the aggregated frame $\overline{A}(t)$ captures it by referencing neighboring independent frames.
The network adds these missing fine details to the independent frame by retracting those details from nearby frames, as shown in the weight map, in order to improve the quality of the final output.
On the other hand, part (b) contains a rapidly moving object, so that the edge of it is blurred when aggregating multiple frames.
Hence, the weight pixels corresponding to the part are large for the independent frame.
Through the compatibility of the independent and aggregated frames, our approach assures high performance.

\begin{figure*}[ht]
\begin{center}
\includegraphics[width=1.0\linewidth]{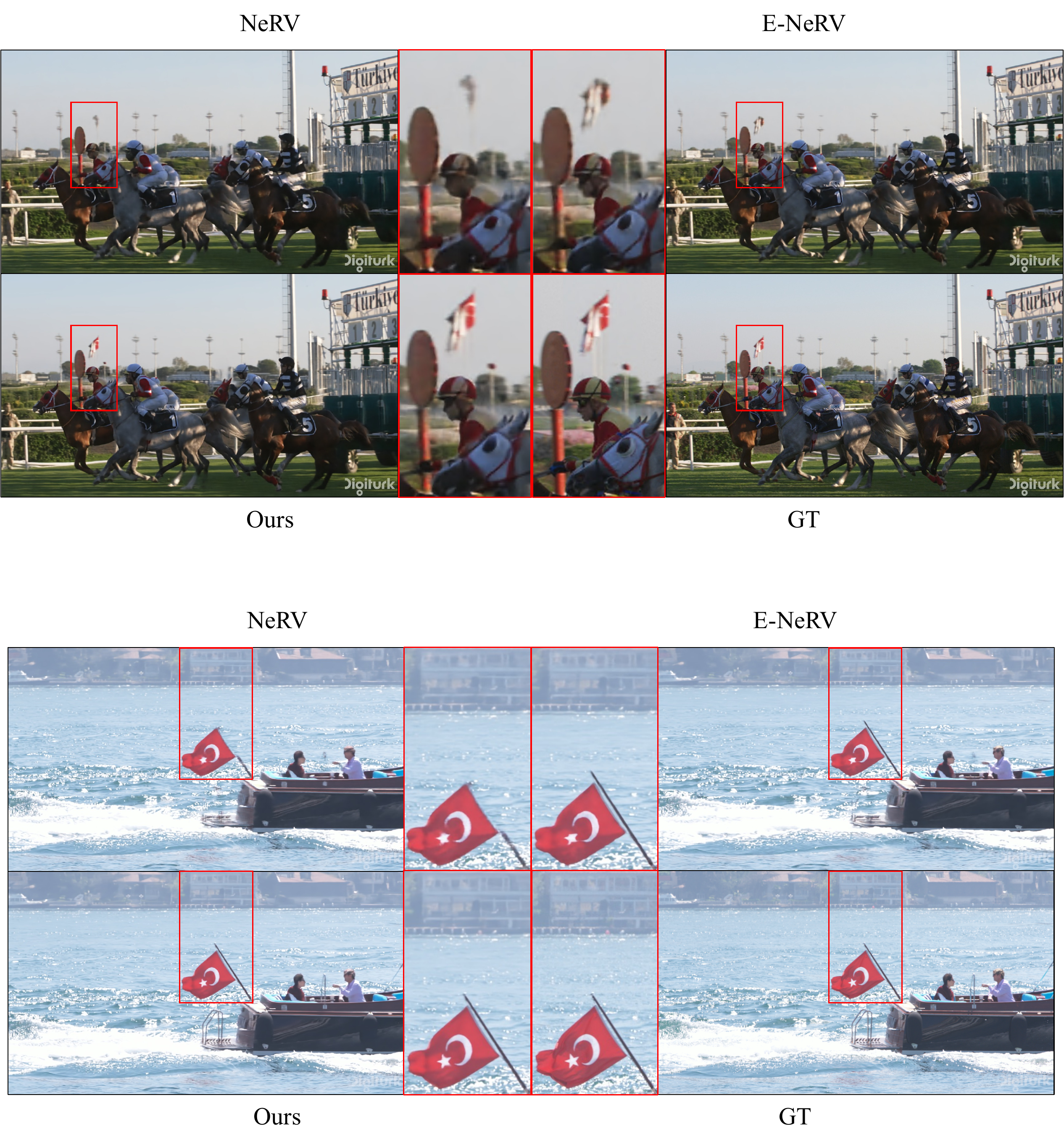}
\end{center}
   \caption{The qualitative results for video representation.}
\label{fig_sup_qual2}
\end{figure*}

\begin{figure*}[ht]
\begin{center}
\includegraphics[width=1.0\linewidth]{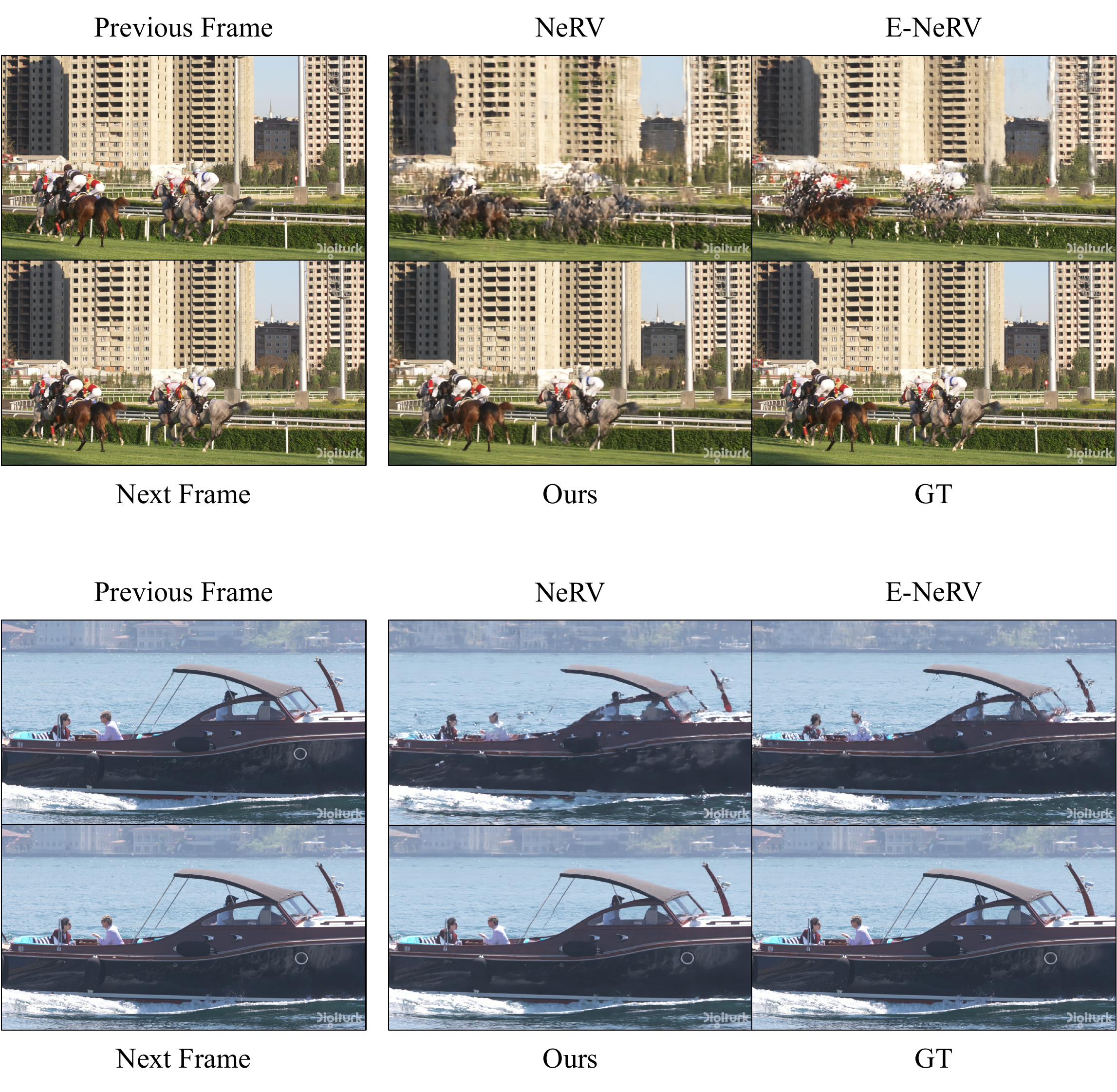}
\end{center}
   \caption{The qualitative results for video frame interpolation.}
\label{fig_sup_qual3}
\end{figure*}

We present the qualitative results of the three similar-sized models; our model, NeRV, and E-NeRV.
Figure \ref{fig_sup_qual2} and \ref{fig_sup_qual3} illustrate the constructed frames for video representation and frame interpolation, respectively.
As shown in those figures, our approach outperforms other frame-wise methods qualitatively, regardless of the task.

\begin{table*}[ht]
\centering
\caption{PSNR and BPP of compression results on UVG videos at various levels.}
\vskip 0.1in
\begin{tabular}{c||c|ccccccc||c}
\hline
($C_1$, $C_2$, S)                                                                                       & \begin{tabular}[c]{@{}c@{}}Video\\ (\#frames)\end{tabular} & \begin{tabular}[c]{@{}c@{}}Beauty\\ (600)\end{tabular} & \begin{tabular}[c]{@{}c@{}}Bospho\\ (600)\end{tabular} & \begin{tabular}[c]{@{}c@{}}Honey\\ (600)\end{tabular} & \begin{tabular}[c]{@{}c@{}}Jockey\\ (600)\end{tabular} & \begin{tabular}[c]{@{}c@{}}Ready\\ (600)\end{tabular} & \begin{tabular}[c]{@{}c@{}}Shake\\ (300)\end{tabular} & \begin{tabular}[c]{@{}c@{}}Yacht\\ (600)\end{tabular} & Avg    \\ \hline\hline
\multirow{2}{*}{\begin{tabular}[c]{@{}c@{}}600: (24, 192, 16)\\ 300: (16, 128, 16)\end{tabular}}  & PSNR                                                       & 33.63                                                  & 34.62                                                  & 38.88                                                 & 33.75                                                  & 26.68                                                 & 33.34                                                 & 28.38                                                 & 32.71  \\
                                                                                                  & BPP                                                        & 0.0187                                                 & 0.0185                                                 & 0.0180                                                & 0.0186                                                 & 0.0187                                                & 0.0148                                                & 0.0186                                                & 0.0182 \\ \hline
\multirow{2}{*}{\begin{tabular}[c]{@{}c@{}}600: (48, 392, 24)\\ 300: (32, 256, 24)\end{tabular}}  & PSNR                                                       & 34.21                                                  & 38.41                                                  & 39.6                                                  & 37.29                                                  & 31.48                                                 & 35.26                                                 & 32.48                                                 & 35.55  \\
                                                                                                  & BPP                                                        & 0.0556                                                 & 0.0551                                                 & 0.0533                                                & 0.0561                                                 & 0.0560                                                & 0.0461                                                & 0.0558                                                & 0.0546 \\ \hline
\multirow{2}{*}{\begin{tabular}[c]{@{}c@{}}600: (64, 512, 32)\\ 300: (48, 384, 32)\end{tabular}}  & PSNR                                                       & 34.52                                                  & 39.88                                                  & 39.72                                                 & 38.41                                                  & 33.64                                                 & 36.75                                                 & 34.41                                                 & 36.76  \\
                                                                                                  & BPP                                                        & 0.0898                                                 & 0.0895                                                 & 0.0870                                                & 0.0909                                                 & 0.0907                                                & 0.0929                                                & 0.0904                                                & 0.0890 \\ \hline
\multirow{2}{*}{\begin{tabular}[c]{@{}c@{}}600: (80, 640, 48)\\ 300: (56, 448, 48)\end{tabular}}  & PSNR                                                       & 34.78                                                  & 40.97                                                  & 39.86                                                 & 39.02                                                  & 35.41                                                 & 37.33                                                 & 35.96                                                 & 37.64  \\
                                                                                                  & BPP                                                        & 0.132                                                  & 0.132                                                  & 0.129                                                 & 0.134                                                  & 0.133                                                 & 0.122                                                 & 0.133                                                 & 0.131  \\ \hline
\multirow{2}{*}{\begin{tabular}[c]{@{}c@{}}600: (96, 768, 48)\\ 300: (64, 512, 48)\end{tabular}}  & PSNR                                                       & 35.05                                                  & 41.73                                                  & 40.01                                                 & 39.41                                                  & 36.72                                                 & 37.81                                                 & 37.3                                                  & 38.33  \\
                                                                                                  & BPP                                                        & 0.182                                                  & 0.181                                                  & 0.178                                                 & 0.184                                                  & 0.183                                                 & 0.154                                                 & 0.183                                                 & 0.180  \\ \hline
\multirow{2}{*}{\begin{tabular}[c]{@{}c@{}}600: (112, 896, 54)\\ 300: (80, 640, 54)\end{tabular}} & PSNR                                                       & 35.35                                                  & 42.32                                                  & 40.18                                                 & 39.71                                                  & 37.78                                                 & 38.52                                                 & 38.34                                                 & 38.91  \\
                                                                                                  & BPP                                                        & 0.240                                                  & 0.240                                                  & 0.234                                                 & 0.242                                                  & 0.241                                                 & 0.231                                                 & 0.241                                                 & 0.239  \\ \hline
\end{tabular}
\label{tab_comp_res}
\end{table*}

\end{document}